# Raisonnement stratifié à base de normes pour inférer les causes dans un corpus textuel

Farid Nouioua
LIPN UMR 7030 du C.N.R.S.
Institut Galilée – Univ. Paris-Nord
93430 Villetaneuse – FRANCE
nouiouaf@lipn.univ-paris13.fr

**Résumé :**

La bonne compréhension d'un texte écrit en langage naturel (*LN*) fait appel à nos connaissances sur les normes du domaine qui permettent d'en déduire des informations implicites bien plus riches que celles exprimées explicitement dans le texte. Ces informations peuvent être remises en cause, mais elles restent admises et utiles tant que le texte ne les contredit pas. Dans ce papier, nous décrivons un système de raisonnement non monotone basé sur les normes du domaine des accidents de la route et permettant d'inférer la cause d'un accident à partir de sa description textuelle. Le lien entre la notion de cause et la notion de norme est établi en supposant que la cause d'un événement anormal est la violation d'une certaine norme du domaine. Les prédicats et les règles du système de raisonnement sont stratifiés, i.e., organisés dans des couches de façon à satisfaire certaines contraintes afin d'obtenir une méthode de raisonnement de complexité raisonnable.

**Mots-clés :**

Normes, Raisonnement Causal, Raisonnement stratifié, sémantique du langage naturel.

## 1. Introduction

Depuis les débuts de l'intelligence artificielle (*IA*), l'une de ses ambitions majeures était la reproduction, sur des ordinateurs, de la capacité humaine de compréhension de textes écrits en langage naturel (*LN*). A l'heure actuelle, Il est admis dans la communauté des chercheurs de ce domaine, du moins avec l'état actuel des choses, que la compréhension automatique universelle du LN est extrêmement complexe voire impossible. Divers niveaux d'analyse interviennent dans la compréhension de la langue :

- Le niveau lexical et le besoin de doter le système de dictionnaires automatiques, avec toutes les questions fondamentales sous-jacentes sur la nature de représentation du sens d'un mot [13]. Ces questions relèvent de la sémantique lexicale, champ de recherche encore très actif en *IA* ;

- les niveaux morphologiques et syntaxiques et le besoin de capter d'une part la variation des mots en fonction du genre, du nombre, du temps …etc. et d'autre part la structure de la phrase et les liens grammaticaux entre ses constituants ;

- le niveau sémantique avec notamment le débat sur l'adéquation de l'approche compositionnelle dans la construction du sens d'une phrase à partir des sens des mots qui la composent [17] ;

- enfin, le niveau pragmatique qui fait appel à des connaissances bien au-delà de celles exprimés explicitement dans le texte afin d'adapter la sémantique d'une phrase au contexte particulier de son énonciation [12]. Ce niveau d'étude ajoute une dimension vague et délicate au processus de compréhension.

Il n'y a pas de limites claires entre ces niveaux qui se chevauchent et interagissent les uns avec les autres pour réaliser la tâche voulue de compréhension.

Ces sérieuses difficultés ont poussé les recherches dans ce domaine à poursuivre des objectifs moins ambitieux mais plus réalistes. Ainsi, on note entre autres le développement des ontologies et leurs multiples applications [3], les outils d'aide à la traduction automatique [5][16], les outils d'extraction automatique de connaissances et de ressources linguistiques à partir de corpus textuels [1], les systèmes de questions réponses [4]…etc.

Ces différents types d'applications sont caractérisés essentiellement par la limitation du domaine d'application d'une part, et la fixation d'un objectif particulier d'étude d'autre part.

Dans le cas de notre travail, l'objectif est de répondre automatiquement à la question de savoir la cause d'un accident à partir d'un texte décrivant ses circonstances en *LN*. Nous travaillons alors sur un domaine bien déterminé qui est celui des accidents de la route. La fixation du domaine d'application et de l'objectif permet de limiter la plage des phénomènes linguistiques qui nous intéressent et de réduire également la difficulté de leurs traitements.

L'accent est mis sur les aspects sémantiques et pragmatiques (on parlera ici de sémantique pour désigner les deux niveaux). On se base sur un résultat très important de l'*IA* dans ce domaine qui montre que la sémantique de la *LN* est de nature inférentielle plutôt que référentielle [6]. Dans cette optique, les inférences jouent un rôle central dans le processus de compréhension et les connaissances explicites exprimées dans un texte ne sont qu'un point de départ d'un processus de raisonnement qui permet de le comprendre. Ceci remet en cause la notion de vérité comme base de la sémantique du *LN*, élément qui caractérise la plupart des théories de la sémantique formelle. Nous défendons par contre l'idée que la connaissance des normes d'un domaine ouvre des possibilités bien plus riches et plus puissantes que la notion de vérité pour comprendre un texte en *LN*. Notre approche consiste donc à centrer la compréhension sur un processus de raisonnement basé sur les normes du domaine d'application, en l'occurrence, celui des accidents de la route pour le présent travail.

## 2. Pourquoi les normes ?

Prenons le texte suivant [8] [9] : *" Le véhicule de devant a freiné soudainement "*. En se basant sur la notion de vérité, nous pouvons déduire de ce texte toutes les conséquences logiques d'une formule du genre :

$$(\exists v,t)\ véhicule(v) \land Instant(t) \land devant(v, \text{'moi'}, t) \land freiner(v, t)$$

Cependant, un lecteur humain est capable d'en déduire plus de conclusions, notamment celles du type :

- *v* et moi circulaient dans la même direction ;
- pas de véhicule entre *v* et moi ;
- je devais freiner lorsque *v* a freiné.

Bien que ces conclusions soient susceptibles d'être remises en cause, elles seront admises tant que rien d'explicite dans le texte ne les contredit. En plus, elles sont indispensables à la bonne compréhension du texte. En fait, l'auteur du texte suppose à priori que le lecteur les inférera en utilisant ses connaissances sur des normes du domaine.

Nous parlons ici d'un sens bien défini du mot "*norme*" qui désigne le déroulement normal et non pas normatif des événements. Ce dernier sens a fait l'objet de nombreux travaux en *IA*, notamment en logique déontique [2].

Les travaux les plus marquants qui se sont basés sur la notion de norme dans la compréhension du *LN* sont ceux sur les *Frames* de Minsky [11] et les *scripts* de Schank [15]. Ces travaux fondés sur des théories issues de la psychologie sont malheureusement informels et fortement liés à leurs domaines d'application initiaux. Leur adaptation à de nouvelles situations est loin d'être triviale.

Nous voulons profiter des progrès réalisés en *IA* autour de la notion de norme, notamment dans le domaine du raisonnement avec le développement des *logiques non monotones*, pour aborder à nouveau le problème avec une approche différente.

## 3. Quelle est notre tâche ?

Les normes sont importantes pour la compréhension du *LN*, mais nous ne disposons pas à l'heure actuelle de supports qui les regroupent pour un domaine donné. L'extraction des normes d'un domaine est une tâche extrêmement utile et permet de faciliter la construction de programmes de compréhension automatique des textes. La difficulté de l'extraction des normes à partir des textes vient, entre autres, du fait que les textes ne les décrivent pas explicitement et se focalisent plutôt sur les déviations des événements par rapport au déroulement normal, c'est-à-dire sur les violations des normes. Notre idée principale consiste alors à extraire les normes d'un domaine en analysant leurs violations dans un texte. Dans un premier temps, on s'intéresse à l'élaboration d'un raisonnement basé sur les normes d'un domaine pour comprendre un texte décrivant un accident de la route en répondant à une question précise : *" Quelle est la cause de l'accident ? "* et en supposant que la cause recherchée est une violation d'une certaine norme (voir plus loin). Le critère de validation est que le système soit capable de donner la même réponse qu'un lecteur ordinaire du texte donne généralement.

Nous utilisons un corpus de (60) textes fourni par la compagnie d'assurance *MAIF*. Chacun de ces textes décrit les circonstances d'un accident. Il est rédigé au verso du constat amiable par les personnes impliquées dans l'accident.

Les textes étudiés sont courts (4 à 5 lignes en moyenne). Ils utilisent un vocabulaire restreint et sont très bien adaptés à notre objectif car les normes sont fortement impliquées pour les comprendre. Ces textes présentent un aspect *argumentatif* : ils sont souvent utilisés par l'auteur pour montrer que c'est l'autre conducteur qui est responsable de l'accident. L'argumentation est en elle-même révélatrice de normes mais, pour simplifier le problème, nous essayons d'ignorer cet aspect dans notre étude en ne nous concentrant que sur les faits déclarés dans le texte et qui sont nécessaires à notre raisonnement.

La question à laquelle nous nous intéressons concerne la *cause*. La notion de *causalité* est une notion très délicate car depuis toujours, il n'y avait pas de consensus sur sa définition exacte. Dans notre travail, nous restons loin des débats philosophiques sur ce sujet et nous cherchons simplement à adopter une vision pratique de la cause qui nous permet de raisonner sur nos textes. Cette vision consiste à privilégier parmi la multitude des causes possibles d'un évènement donné, une action volontaire dont l'exécution conduit à un futur différent de celui qui en résulterai si l'action n'avait pas eu lieu [7]. C'est un point de vue particulier de l'*I.A* sur la causalité. En fait, cette conception de la cause est très utile pour l'*I.A* dans des applications pratiques telles que le diagnostic et la prédiction. Pour établir un lien entre la notion de cause et celle de norme, nous nous appuyons sur l'hypothèse que la cause d'un événement normal est la norme elle-même, alors que la cause d'un événement anormal est la violation d'une norme que l'on appelle ici anomalie [8]. Parmi les différentes violations de normes possibles dans un texte, nous cherchons à distinguer celle qui peut être considérée comme la cause la plus plausible de l'accident et que l'on appelle " *Vraie anomalie* " des autres violations qui résultent de la première et qui sont appelées " *Fausses anomalies* ".

**4. Langage de représentation**

Pour raisonner sur les normes, il nous faut d'abord un langage adéquat pour leur représentation. Le langage proposé permet d'exprimer des *propriétés* et des *modalités* pour des agents en tenant compte de l'aspect *temporel*. Il est également adapté à un raisonnement non monotone. Les aspects considérés dans notre langage ont été abondamment traités par des logiques et des formalismes adaptés. Cependant les raisonnements dont on a besoin sont très rudimentaires, alors que les logiques qui traitent ces questions sont censées répondre, au prix d'une complexité qui serait ici injustifiée, à des tas de questions qu'on ne se posera jamais. Notre stratégie consiste donc à utiliser d'une part des prédicats de premier ordre réifiés capables d'exprimer les aspects nécessaires au raisonnement et d'autre part des règles d'inférences strictes ou non monotones pour effectuer ce raisonnement. En cas de besoin, les axiomes nécessaires sont exprimées explicitement par des règles d'inférences. Les détails du langage que nous avons proposé sont décrits dans [9]. Nous présentons ici un bref aperçu sur ces éléments de base.

**4.1. Une logique de premier ordre réifiée**

Afin de quantifier et de raisonner sur des noms de propriétés tout en restant dans le cadre d'une logique de premier ordre, nous utilisons la technique de *réification*, communément utilisée en *IA*. Un prédicat binaire *P(x, y)* est écrit *Vrai(P, x, y)*.

Nous utilisons la fonction unaire *neg* pour obtenir à partir d'une propriété sa négation, avec :

$$Vrai(neg(P), A, t) \leftrightarrow \neg Vrai(P, A, t)$$

Les prédicats d'arité supérieure à 2, sont représentés en utilisant une fonction binaire *Combine*, qui permet de construire une nouvelle propriété complexe en combinant une propriété et un autre argument. Par exemple, un prédicat ternaire *Q(x, y, z)* est représenté par : *Vrai(Combine(Q, x), y, z)*.

**4.2. Le temps**

L'ordre de production des évènements décrits dans un texte est important pour la détermination des anomalies. Notre modèle du temps est linéaire, nous nous intéressons uniquement aux événements effectivement réalisés. Le temps avec branchement et futurs multiples [10] bien que très important, ne nécessite pas une représentation explicite dans notre modèle. Nous nous contentons d'une représentation implicite à l'aide des modalités (Voir plus loin). La scène est décomposée en une suite d'intervalles successifs numérotés. Chaque intervalle est caractérisé par un ensemble de littéraux gardant une valeur de vérité stable.

Les prédicats de notre langage sont de la forme *Vrai(P, A, t)*, où *P* est une propriété simple ou complexe (exprimée par *Combine*), *A* est un agent et *t* est un entier naturel qui désigne le numéro de l'intervalle de temps dans lequel *P* est vérifiée. La fonction *Combine* n'est en général pas associative. Pour décider dans le cas d'une propriété ternaire quel paramètre sera affecté à la fonction *Combine*, nous nous appuyons sur la forme général du prédicat *Vrai*. A titre d'exemple, prenons le fait que le véhicule *A* a heurté le véhicule *B* à un instant *T*. Le problème est de choisir entre *A* et *B* le paramètre qui sera affecté à *Combine*. Puisque l'agent de l'action *Heurter* est le véhicule A, nous le prenons comme deuxième paramètre du prédicat *Vrai*. La fonction *Combine* permet de construire la propriété complexe : ″Heurter le véhicule B″ à partir de la propriété simple *Heurter* et du paramètre B. Le prédicat qui en résulte est : *Vrai(Combine(Heurter, A), B, T)*.

*Exemples*

    *Vrai(Arrêt, V, 2)* : le véhicule *V* est à l'arrêt durant l'intervalle de temps *2*.

    *Vrai(Combine(Suiv, W), V,1)* : le véhicule *V* suit le véhicule *W* durant l'intervalle de temps *1* (par abus du langage on dit « à l'instant *t* » au lieu de « durant l'intervalle de temps *t* »)

### 4.3. Les modalités

Certaines propriétés du texte s'expriment plus aisément par des modalités. La technique de réification nous permet également de représenter les modalités dans le cadre d'une logique des prédicats du premier ordre (*LPPO*). Nous utilisons dans notre travail deux types de modalités : la première est une modalité de devoir qui est une sorte de nécessité; le prédicat *Doit(P, A, t)* veut dire qu'à l'instant *t*, l'agent *A* doit atteindre la propriété *P*. La deuxième est une modalité de capacité. Il s'agit d'une sorte de possibilité; le prédicat *En_Mesure(P, A, t)* exprime le fait qu'à l'instant *t* l'agent *A* est capable d'atteindre la propriété *P*.

### 4.4. La non monotonie

Afin de tenir compte de l'aspect non monotone du raisonnement sur les normes, nous utilisons en plus des implications strictes de la forme $A \rightarrow B$, des défauts normaux et semi-normaux [14]. Pour abréger l'écriture, un défaut normal de la forme $\dfrac{A : B}{B}$ est noté A : B, et un défaut semi-normal $\dfrac{A : B \wedge C}{B}$ est noté *A : B[C]*.

### 4.5. Représentation formelle d'une anomalie

Les deux formes possibles d'une vraie anomalie sont :

    *Doit(P, A, t) ∧ En_Mesure(P, A, t) ∧ Vrai(P', A, t+1) ∧ Incompatible(P, P') → Vrai_Anomalie*

Si à l'instant *t*, *A* doit atteindre *P* et s'il est en mesure de le faire, et qu'à l'instant *t+1*, une propriété *P'* incompatible avec *P* est vraie, alors il y a vraie anomalie.

    *Vrai(Combine(Cause_Perturb_Anormale, X), A, t) → Vraie_Anomalie.*

Si *X* représente une cause de perturbation anormale pour *A* à l'instant *t*, alors il y a vraie anomalie.

La forme d'une fausse anomalie est la suivante :

    *Doit(P, A, t) ∧ ¬En_Mesure(P, A, t) ∧ Vrai(P', A, t+1) ∧ Incompatible(P, P') → Fausse_Anomalie*

(cette formule ne diffère de la première que par le fait que l'agent *A* n'est pas en mesure de faire à l'instant *t* ce qu'il devrait faire).

## 5. Architecture générale du système

La figure 1. montre l'architecture générale de notre système. L'accent est mis dans ce papier sur la dernière étape (la partie droite de l'architecture) qui permet de passer des prédicats sémantiques aux prédicats du noyau au moyen d'un raisonnement sémantique. La nature du raisonnement sémantique sera discutée dans la section suivante. Dans cette section, nous donnons une brève description des différentes étapes de l'architecture.

### 5.1. Analyse syntaxique

La première étape du processus consiste à effectuer une *analyse syntaxique* du texte. Le résultat attendu de cette étape est un ensemble de *prédicats linguistiques*. Il s'agit d'un petit nombre de relations syntaxiques entre les mots du texte. Les prédicats linguistiques que l'on cherche à extraire à partir du texte sont :

    *Sujet(V, N), Objet(V, N)* : le nom *N* est le sujet (resp. l'objet) du verbe *V* ;

*Qualif-N(N, A), Qualif-V(V, A)* : *A* est une qualification du nom *N* (resp. du verbe *V*). C'est le cas entre autres des adjectives et des adverbes ;

*Compl-N(X, N, C), Compl-V(X, V, C)* : C est un complément du nom *N* (resp. du verbe *V*) introduit par le mot *X* ;

*Support(X, Y)* : *X* est un support de *Y*. Par exemple dans *"A est venu heurter B"*, nous disposons de la relation *Support(venir, heurter)*.

### 5.2. Raisonnement linguistique

L'objectif de cette étape du raisonnement est de transformer les prédicats linguistiques en des *prédicats sémantiques*. Ces derniers sont censés représenter l'essentiel du contenu sémantique du texte. Le *raisonnement linguistique* est basé sur des connaissances d'ordre linguistique (étude lexicale des mots du domaine, connaissances syntaxiques, …). Nous avons dressé une liste d'un nombre raisonnable (environ 50) de prédicats sémantiques capables de véhiculer l'information explicite d'un texte et de déclencher le raisonnement sémantique basé sur les normes à l'étape suivante. Un échantillon des prédicats sémantiques est présenté dans l'annexe 1.

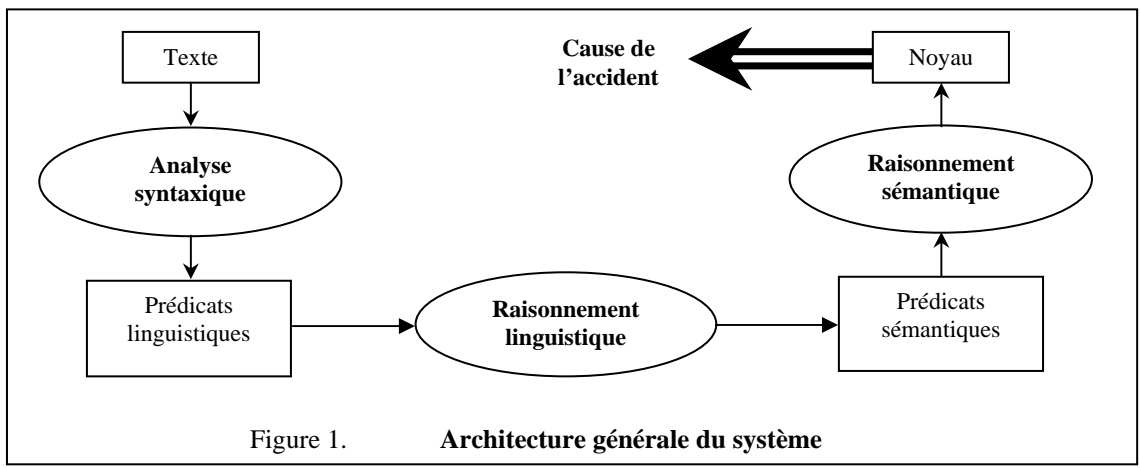

Figure 1.     **Architecture générale du système**

### 5.3. Raisonnement sémantique

Le *raisonnement sémantique* est un raisonnement non monotone qui permet d'enrichir les prédicats sémantiques résultant de l'étape précédente par de nouvelles connaissances implicites. Il est effectué à l'aide d'un ensemble de règles strictes et de défauts. La conception des règles s'appuie sur nos connaissances des normes du domaine de la route. Les inférences effectuées au niveau du raisonnement sémantique permettent la convergence des conclusions vers un ensemble restreint de 6 prédicats constituant ce qu'on appelle le noyau (Voir figure 1.). Les prédicats du noyau sont en mesure d'exprimer n'importe quelle anomalie évoquée dans un texte. Le raisonnement sémantique se termine dès qu'il arrive à détecter l'une des deux formes d'une vraie anomalie au moyen des prédicats du noyau, à savoir :

*Vrai(Arrêt, A, t)* ; Le véhicule *A* est à l'arrêt à l'instant *t* ;

*Vrai(Contrôle, A, t)* ; Le véhicule *A* est contrôlé par son conducteur à l'instant *t* ;

*Vrai(Roule_Assez_Lentement, A, t)* ; Le véhicule *A* roule assez lentement à l'instant *t* ;

*Vrai(Démarrer, A, t)* ; Le véhicule *A* démarre à l'instant *t* ;

*Vrai(Reculer, A, t)* ; Le véhicule *A* recule à l'instant *t* ;

*Vrai(Combine(Cause_Perturb_Anormale, X), A, t)* ; *X* est une cause de perturbation anormale pour *A* à l'instant *t*.

Nous avons conçu à l'heure actuelle plus de 100 règles sémantiques qui assurent la couverture d'une partie importante du corpus. Les quelques exemples suivants montrent la nature des règles utilisées :

- Si un véhicule est stationné à l'instant t, alors il est à l'arrêt à cet instant.

$$Vrai\ (Stationné, A, t) \rightarrow Vrai\ (Arrêt, A, t)$$

- En général, si les véhicules *A* et *B* circulent dans une même file à l'instant *t* et que le véhicule *B* heurte le véhicule *A* à cet instant, alors on peut inférer que *B* suivait *A* à l'instant *t-1*.

*Vrai(Combine(Même_File, A), B, t) ∧ Vrai(Combine(Heurter, A), B, t) : Vrai(Combine(Suiv, A), B, t-1)*

- Si à l'instant *t*, le véhicule *A* suit le véhicule *B* dans une file, et *B* roule assez lentement, alors à cet instant, *A* doit rouler assez lentement.

*Vrai(Combine(Suiv, B), A, t) ∧ Vrai(Roule_Lentement, B, t) → Doit(Roule_Lentement,A, t)*

- En général, si le véhicule *B* démarre à l'instant *t-1*, et qu'il y a un choc à l'instant *t* entre *B* et un autre véhicule *A*, alors à l'instant *t-2 B* avait le devoir de ne pas démarrer. Cette inférence est bloquée si, à l'instant *t-1*, *B* est prioritaire sur *A* ou *A* suit *B* dans une file.

*Vrai(Démarrer, B, t-1) ∧ Vrai(Combine(Choc, A), B, t) : Doit(non(Démarrer), B, t-2)*
*[¬Vrai(Combine(Prioritaire, A), B, t-1), ¬Vrai(Combine(Suiv, B), A, t-1)].*

## 6. Une approche de raisonnement : la stratification

### 6.1. La méthode

Effectuer le raisonnement sémantique revient à calculer l'extension d'une théorie exprimée en logique des défauts *<W, D>* ; *W* contient l'ensemble des règles strictes de notre système ainsi que l'ensemble des littéraux provenant du texte et résultant du raisonnement linguistique. Ces littéraux sont appelés faits. *D* est l'ensemble des défauts normaux et semi normaux du système.

En général, le calcul d'extensions d'une telle théorie est indécidable. Pour surmonter cette difficulté, nous imposons certaines contraintes aux règles afin de réduire l'espace de recherche et obtenir un algorithme de complexité raisonnable. La technique que nous utilisons dans notre travail consiste à stratifier les prédicats et les règles du système. i.e., Les prédicats et les règles du système appartiennent à des couches ordonnées et numérotées de *N* (la plus externe) à *1* (la plus interne). Le raisonnement se fait progressivement couche par couche. Un prédicat *P* appartient à la couche *i* si et seulement si sa valeur de vérité est décidée au plus tard lorsqu'on raisonne sur la couche *i*. C'est-à-dire que les couches *j* (*j < i*) n'agissent pas sur la valeur de vérité de *P*.

Plus formellement, pour un prédicat ou une règle donnée *PR*, *Couche(PR) = i* si et seulement si *i* est le numéro de la couche à laquelle appartient *PR*. La répartition des prédicats sur les couches doit vérifier les 3 propriétés suivantes :

(1) Chaque règle stricte *(R)* de la forme *A → B* est telle que :

- *A* et *B* sont des conjonctions de littéraux :
  $A = A_1 \wedge A_2 \ldots A_m$ et $B = B_1 \wedge B_2 \wedge B_p$ (*m, p > 0*);
- $Min(couche(A_i)) > Max(couche(B_j))$.

Cette règle appartient à la couche de numéro : $Max(couche(B_j))$. On aura donc :

$$Couche(R) < Min(Couche(A_i))$$

(2) Chaque défaut normal *(DN)* de la forme *A : B* est tel que :

- *A* est une conjonction de littéraux :
  $A = A_1 \wedge A_2 \ldots A_m$ (*m > 0*) ;
- *B* est un littéral ;
- $Min(couche(A_i)) > couche(B)$.

Ce défaut appartient à la couche de numéro : *couche(B)*. On aura donc :

$$Couche(DN) < Min(Couche(A_i))$$

(3) Chaque défaut semi-normal *(DSN)* de la forme *A : B[C]* est tel que :

- *A* et *C* sont des conjonction de littéraux :
  $A = A_1 \wedge A_2 \ldots A_m$ et $C = C_1 \wedge C_2 \ldots C_q$ (*m, p > 0*) ;
- $Min(couche(A_i)) > couche(B)$ ;
- $Min(couche(C_k)) > couche(B)$.

Ce défaut appartient à la couche de numéro *couche(B)*. On aura donc :

$$Couche(DSN) < Min(couche(C_k), couche(A_i))$$

### 6.2. Modélisation par un graphe

Les prédicats qui composent les littéraux des prémisses, des conclusions et des justifications des règles représentent les sommets du graphe. Les arcs expriment les relations de dépendance entre les prédicats. Ils sont déduits à partir des règles comme suit :

- Pour toute règle *(R)* de la forme $A \rightarrow B$ et pour tout couple de prédicats *P* et *Q* tel que *P* ou $\neg P$ figure parmi les littéraux de *A*, et *Q* ou $\neg Q$ figure parmi les littéraux de *B*, on crée un arc de *P* vers *Q* ;

- Pour tout défaut normal *(DN)* de la forme *A : B*, si $B_1$ est le prédicat correspondant au littéral *B* (*B* = $B_1$ ou *B* = $\neg B_1$), alors pour tout prédicat *P* tel que *P* ou $\neg P$ figure parmi les littéraux de *A*, créer un arc de *P* vers $B_1$ ;

- Pour tout défaut semi normal *(DSN)* de la forme *A : B[C]*, si $B_1$ est le prédicat correspondant au littéral *B* (*B* = $B_1$ ou *B* = $\neg B_1$), alors pour tout couple de prédicats *P*, *Q* tel que *P* ou $\neg P$ figure parmi les littéraux de *A*, et *Q* ou $\neg Q$ figure parmi les littéraux de *C*, créer un arc de *P* vers $B_1$ et un arc de *Q* vers $B_1$.

*Exemples*

1. Le graphe correspondant aux règles

   $A \rightarrow B$, *A : ¬D*, *D : K [A]*

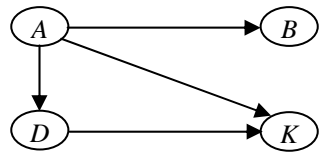

2. Le graphe correspondant aux règles

   $A \rightarrow B$, *B : C*, *X : A [C]*

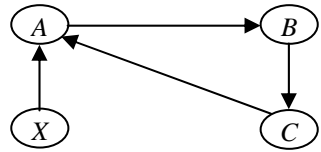

*Définition*

Une théorie *<W, D>* est stratifiée si et seulement si le graphe qui la modélise est un graphe sans cycle.

*Corollaire*

Les couches d'une théorie *<W, D>* stratifiée correspondent aux niveaux du graphe sans cycle qui en résulte.

### 7. Cas de notre système

Dans notre système, nous disposons de deux niveaux de stratification. Le premier niveau contient quatre couches globales. Cette première décomposition est effectuée en se basant sur les modalités. L'idée est que le sens du raisonnement n'est pas arbitraire : en gros, une première étape consiste à enrichir le système avec de nouveaux faits avant de passer à l'inférence des devoirs, puis des capacités pour finir avec la détermination des anomalies. Dans le deuxième niveau, nous décomposons les deux premières couches en sous couches selon les propriétés sur lesquelles portent les prédicats. Nous pouvons voir cette décomposition comme ayant un seul niveau dans lequel les sous couches de la première couche précèdent celles de la deuxième qui à leur tour, précèdent les deux dernières couches. Dans ce qui suit, nous décrirons brièvement la structure de nos couches et sous couches par rapport à l'ensemble de règles qui ont été développées actuellement et qui couvrent une partie significative du corpus.

- La première couche regroupe les prédicats ayant la modalité vide, c'est-à-dire, dans lesquels seules les instances du prédicat *Vrai(P, A, t)* sont utilisées. Les prémisses et les conclusions des règles de cette couche ne contiennent donc que ce prédicat. A l'intérieur de cette couche, nous avons effectué un deuxième niveau de stratification. Nous avons obtenu 10 sous couches permettant de satisfaire les contraintes de stratification (Voir l'annexe 2). ;

- La seconde couche est formée des prédicats de devoir. C'est-à-dire les instances de la forme *Doit(P, A, t)*. Les règles de cette couche permettent d'inférer les devoirs des agents en se basant sur les faits extraits du texte et enrichis à l'aide des règles de la première couche. Cette couche est décomposée à son tour en deux sous couches (voir annexe 2) ;

- La troisième couche permet d'inférer des informations sur les capacités des agents à accomplir leurs devoirs. Ces informations sont exprimées à l'aide du prédicat *En_Mesure(P, A, t)* ;
- La quatrième couche est formée des trois règles qui permettent d'identifier les vraies et les fausses anomalies (voir section 4.5.).

## 8. Exemple

Prenons l'exemple du texte suivant extrait de notre corpus :

*"Etant à l'arrêt au feu rouge, j'ai été percuté à l'arrière par le véhicule B, son conducteur n'ayant pas réussi à s'arrêter".*

Les prédicats sémantiques tirés de ce texte sont :

*Vrai(Arrêt, 'V', 1), Vrai(Combine(Feu, Rouge), 'V', 1),*
*Vrai(Combine(Heurter, 'V'), 'W', 2), Vrai(Combine(Position_Choc, arrière), V, 2), ¬Vrai(Arrêt, 'W', 2)*

*'V'* et *'W'* désignent respectivement le véhicule de l'auteur du texte et celui désigné par *"véhicule B"* dans le texte.

Comme l'indique le paramètre temporel dans les prédicats, la scène de l'accident est décomposée en deux intervalles de temps.

Dans le raisonnement effectué au niveau de la première couche, huit sous-couches (parmi les dix existantes) sont impliquées. Les ensembles des propriétés que regroupent ces sous-couches sont :

*{Heurter, Position_Choc}, {Choc}, {Arrêt, Eviter, Obstacle}, {Contrôle}, {Prévisible}, {Suiv},*
*{Cause_Arrêt}, {Cause_Arrêt_Ulterieur}*

Parmi les inférences effectuées à ce niveau, citons les inférences utiles suivantes :

- Si *B* heurte *A* à l'instant *t*, alors il y a un choc entre *A* et *B* à cet instant *t* :

*Vrai(Combine(Heurter, A), B), t) → Vrai(Combine(Choc, A), B, t)*   (sous-couche 2)

Cette règle permet d'inférer le littéral *Vrai(Combine(Choc, 'V'), 'W', 2)* (avec *A = 'V', B = 'W' et t = 2*).

- Généralement, si un choc se produit entre *A* et *B* à l'instant *t* et que la position du choc *A* est son arrière, alors *B* suivait *A* à l'instant *t-1* précédant le choc. Cette inférence est inhibée si on arrive à déduire que *B* n'avait pas le contrôle à l'instant *t-1*.

*Vrai(Combine(Choc, A), B, t) ∧ Vrai(Combine(Position_Choc, Arrière), A, t) :*
*Vrai(Combine(Suiv, A), B, t-1)[Vrai(Contrôle, B, t-1)]*

Ce défaut semi-normal permet d'inférer le littéral *Vrai(Combine(Suiv, 'V'), 'W', 1)* (avec *A = 'V', B = 'W', t = 2*)

La règle suivante est un exemple qui illustre l'inférence des devoirs des agents :

- Si *A* est le suivant de *B* (dans une file) et que *B* s'arrête, alors *A* doit s'arrêter :

*Vrai(Combine(Suiv, B),A, t) et Vrai(Arrêt, B, t) → Doit(Arrêt,A, t)*

Cette règle permet d'inférer le devoir de *'W'* à s'arrêter à l'instant *1* : *Doit(Arrêt, B, 1)* (avec *A = ' W', B = 'V' et t = 1*)

Les inférences de la troisième couche permettent d'en déduire les capacités des agents. Dans notre exemple, nous nous intéressons en particulier de la capacité de l'agent *'W'* à s'arrêter à l'instant *1*. La règle principale qui permet de définir les capacités des agents est :

*En_Mesure(P, A, t) ↔ (∃Act) Action(Act) ∧ Cause_Pot(P, Act) ∧ Disponible(Act, P, A, t)*

La sémantique de cette règle est que l'agent *A* est en mesure d'atteindre la propriété *P* à l'instant *t* si et seulement s'il existe une action *Act* qui lui est disponible à l'instant *t* et qui représente une cause potentielle pour la propriété *P*.

Le lecteur peut trouver plus de détails sur cette formule dans [9]. Nous précisons ici juste que nous disposons de deux tables *Action* et *Cause_Pot* qui regroupent (statiquement) les occurrences des prédicats correspondants. Nous recensons dans notre système un nombre limité d'actions. Le prédicat *Cause_Pot(P, Act)* exprime le fait que, lorsqu'on observe un effet *P*, on pense directement à ce que l'action *Act* a été exécutée et inversement, si l'action *Act* est exécutée, on s'attend naturellement à ce que l'effet *P* se produise.

Le prédicat *Disponible(Act, P, A, t)* exprime le fait qu'à l'instant *t*, l'agent *A* est capable d'exécuter l'action *Act* et qu'il n'y a pas de circonstances particulières qui inhibent la relation cause-effet entre l'exécution de l'action *Act* et l'obtention de son effet *P*.

Nous arrivons au moyen de cette règle à inférer la capacité de '*W*' à s'arrêter à l'instant *1* : *En_Mesure(Arrêt, 'W', 1)* (avec *P = Arrêt, A = 'W', t = 1* et *Act = Freiner*).

Par ailleurs, en appliquant les deux règles (pour *P = Arrêt, A = 'W'* et *t = 2*) :

$$(\forall P, A, t)\ Vrai(neg(P), A, t) \leftrightarrow \neg Vrai(P, A, t)$$
$$(\forall P)\ Incompatible(P, neg(P))$$

Nous obtiendrons les prédicats *Vrai(neg(Arrêt), 'W', 2)* et *Incompatible(Arrêt, neg(Arrêt))*.

A ce stade du raisonnement tout est prêt pour appliquer la règle d'inférence d'une anomalie (couche 4) présentée plus haut :

$$Doit(P, A, t) \land En\_Mesure(P, A, t) \land Vrai(P', A, t) \land Incompatible(P, P') \rightarrow Vraie\_Anomalie$$

La cause détectée peut être alors exprimée par :

" *Le véhicule 'W' ne s'est pas arrêté à un moment ou il devait le faire* " .

## 9. Conclusion et perspectives

Dans ce papier, nous avons présenté une approche de raisonnement non monotone basée sur les normes et permettant d'inférer la cause d'un accident à partir de sa description textuelle. Nous nous sommes appuyé sur une hypothèse qui consiste à voir la cause d'un événement anormal comme étant la violation d'une certaine norme du domaine. La non monotonie de notre raisonnement est exprimée à l'aide de la logique des défauts de Reiter en utilisant des défauts normaux et semi-normaux. Pour surmonter le problème de l'indécidabilité du calcul des extensions d'une théorie des défauts dans le cas général, nous avons proposé une méthode de stratification des prédicats et des règles du système avec une heuristique qui consiste à saturer les couches au fur et à mesure de l'avancement du raisonnement jusqu'à l'obtention de la forme recherchée d'une vraie anomalie.

Une implémentation du système de raisonnement décrit est en cours. De multiples points restent à développer dans le futur, particulièrement :

- Compléter l'implémentation du système de raisonnement afin de valider nos hypothèses sur l'ensemble des textes du corpus.
- Tester le système sur de nouveaux textes décrivant des accidents de la route.
- Généraliser l'approche sur d'autres domaines.

**Annexe1.** Echantillon des prédicats sémantiques

| | |
|---|---|
| *Vrai(Abord_Carrefour, A, t)* | *A* est à l'abord d'un carrefour |
| *Vrai(Combine(A_gauche, B), A, t)* | *A* est à gauche de *B* |
| *Vrai(Avec_Conducteur, A, t)* | Le véhicule *A* est avec conducteur |
| *Vrai(Combine(Cause_Non_contrôle, X), A, t)* | *X* est une cause de perte de contrôle pour *A* |
| *Vrai(Contrôle , A, t)* | *A* possède le contrôle |
| *Vrai(Démarrer, A, t)* | *A* démarre |
| *Vrai(Combine(Feu, X), A, t)* | *A* est devant un feu de circulation de couleur *X* |
| *Vrai(Combine(Heurter, B), A, t)* | *A* heurte *B* |
| *Vrai(Combine(Position_choc, X), A, t)* | La position du choc reçu par *A* est *X* |
| *Vrai(Combine(Prioritaire, B), A, t)* | *A* est prioritaire sur *B* |
| *Vrai(Combine(Sens_Inverse, A), B, t)* | *A* et *B* circulent dans des sens inverses |
| *Vrai(Stationné , A, t)* | *A* est stationné |
| *Vrai(Stop, A, t)* | Il exist un panneau stop que *A* doit respecter |
| *Vrai(Combine(Suiv, B), A, t)* | *A* est le suivant de *B* dans une file |

**Annexe 2.** Les sous couches des couches 1 et 2.

### Couche 1

*Sous-couche 1.* *{Position_Choc, Arrive_En_File, Stop, Police, Cedez_Passage, Piéton, Virage, Stationné, Prioritaire, Erreur_Commande, Manque_Visibilité, Feu, Heurter, Véhicule, Avec_Conducteur, Frein_Anormal, Reculer}*
*Sous-couche 2.* *{Choc, Démarrer}*
*Sous-couche 3.* *{Eviter, Arrêt, Obstacle}*
*Sous-couche 4.* *{Cause_Non_Contrôle}*
*Sous-couche 5.* *{Contrôle}*
*Sous-couche 6.* *{Prévisible}*
*Sous-couche 7.* *{Roule_Assez_Lentement, Cause_Perturb_Anormal}*
*Sous-couche 8.* *{Même_File, Suiv}*
*Sous-couche 9.* *{Cause_Arrêt}*
*Sous-couche 10.* *{Cause_Arrêt_Ultérieur}*

### Couche 2

*Sous-couche 1.* *{Doit(Contrôle, A, t), Doit(Avec_Conducteur, A, t), Doit(Combine(Eviter, A), B, t), Doit(non(Démarrer), B, t-2), Doit(non(Reculer), B, t-2), Doit(Roule_Lentement, A, t)}*
*Sous-couche 2.* *{Doit(Arrêt,A,t)}*